# Kinematics analysis of the parallel module of the VERNE machine

D. Kanaan[*]    Ph. Wenger[†]    D. Chablat[‡]

Institut de Recherche en Communications et Cybernétique de Nantes,
UMR CNRS 6597, 1 rue de la Noë, 44321 Nantes, France

**Abstract**—*The paper derives the inverse and forward kinematic equations of a spatial three-degree-of-freedom parallel mechanism, which is the parallel module of a hybrid serial-parallel 5-axis machine tool. This parallel mechanism consists of a moving platform that is connected to a fixed base by three non-identical legs. Each leg is made up of one prismatic and two pair spherical joint, which are connected in a way that the combined effects of the three legs lead to an over-constrained mechanism with complex motion. This motion is defined as a simultaneous combination of rotation and translation.*

## I. Introduction

Parallel kinematic machines (PKM) are known for their high structural rigidity, better payload-to-weight ratio, high dynamic performances and high accuracy [1], [2], [3]. Thus, they are prudently considered as attractive alternatives designs for demanding tasks such as high-speed machining [4].

Most of the existing PKM can be classified into two main families. The PKM of the first family have fixed foot and variable–length struts, while the PKM of the second family have fixed length struts with moveable foot points gliding on fixed linear joints [5].

In the first family, we distinguish between PKM with six degrees of freedom generally called Hexapods and PKM with three degrees of freedom called Tripods [6], [7].

Hexapods have a Stewart–Gough parallel kinematic architecture. Many prototypes and commercial hexapod PKM already exist, including the VARIAX (Gidding and Lewis), the TORNADO 2000 (Hexel).

We can also find hybrid architectures like the TRICEPT machine from Neos-robotics [8] which is composed of a two-axis wrist mounted in series to a 3-DOF "tripod" positioning structure.

In the second family, we find the HEXAGLIDE (ETH Zürich) which features six parallel and coplanar linear joints. The HexaM (Toyoda) is another example with three pairs of adjacent linear joints lying on a vertical cone [9]. A hybrid parallel/kinematic PKM with three inclined linear joints and a two-axis wrist is the GEORGE V (IFW Uni Hanover).

Many three-axis translational PKMs belong to this second family and use architecture close to the linear Delta robot originally designed by Clavel for pick-and-place operations [10]. The Urane SX (Renault Automation) and the QUICKSTEP (Krause and Mauser) have three non-coplanar horizontal linear joints [11].

Many researches have made contributions to the study of the kinematics of these PKMs. Most of these articles focused on the discussion of both the analytical and numerical methods [12], [13].

The purpose of this paper is to formulate analytic expressions in order to find all possible solutions for the inverse and forward kinematics problem of the VERNE machine. Then we identify these solutions in order to find the solution that satisfies the end-user.

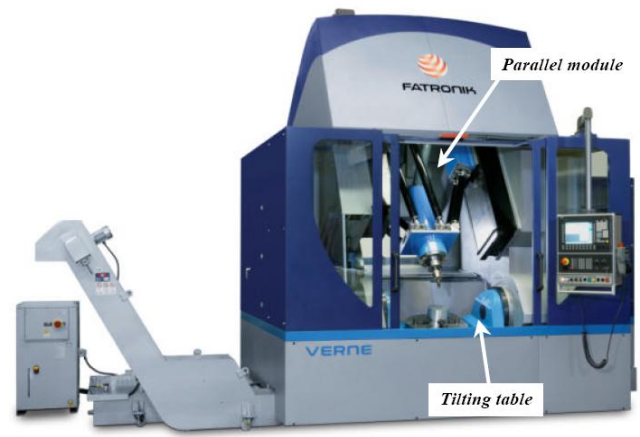

Fig. 1. Overall view of the VERNE machine

The VERNE machine is a 5-axis machine-tool that was designed by Fatronik for IRCCyN [14], [15]. This machine-tool consists of a parallel module and a tilting table as shown in figure 1. The parallel module moves the spindle mostly in translation while the tilting table is used to rotate the workpiece about two orthogonal axes.

In the following three sections, we present the VERNE parallel module, its geometric equations and mobility analysis. Symbolic solutions of the inverse kinematics problem are reported in Section V. Section VI is devoted to the resolution of the forward kinematics problem. Finally a conclusion is given in Section VII.

---
[*] E-mail: daniel.kanaan@irccyn.ec-nantes.fr
[†] E-mail: philippe.wenger@irccyn.ec-nantes.fr
[‡] E-mail: damien.chablat@irccyn.ec-nantes.fr

## II. Description and mobility of the parallel module

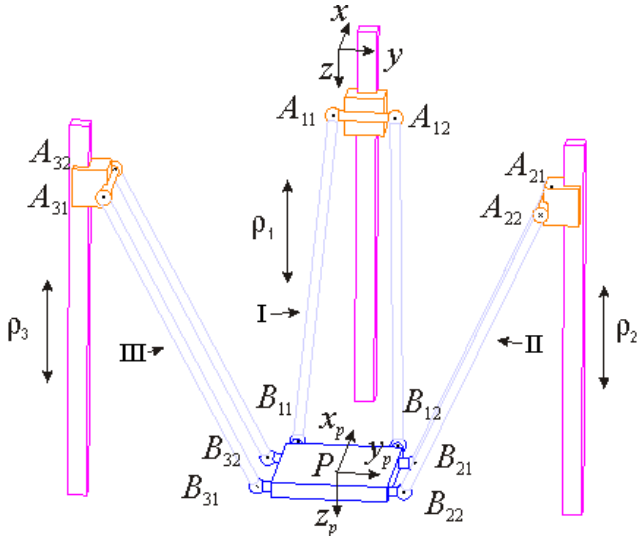

Fig. 2. Schematic representation of the parallel module

Figure 2 shows a scheme of the parallel module of the VERNE machine. The kinematic architecture can be described by a simple scheme shown in figure 3, where joints are represented by rectangles and links between those joints are represented by lines (P and S indicate prismatic and spherical joint, respectively). The moving platform is rectangular. The vertices of this platform are connected to a fixed-base plate through three legs I, II and III. Each leg uses pairs of rods linking a prismatic joint to the moving platform through two pair spherical joints. Legs II and III are two identical parallelograms. Leg I differs from the other two legs in that $A_{11}A_{12} \neq B_{11}B_{12}$, that is, it is not an articulated parallelograms. The movement of the moving platform is generated by the slide of three actuators along three vertical guideways.

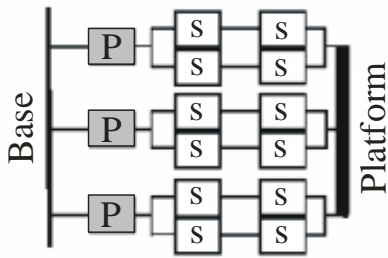

Fig. 3. Joints and loops graph of VERNE

Using the Grubler formula recalled in Equation (1), it can be proved that the mobility $m$ of the platform is equal to three:

$$m = 6(N_p - N_t - 1) + \sum_{i=1}^{N_t} f_i - m_{\text{int}} \quad (1)$$

where $m$ denotes the mobility of the manipulator, $N_p$ is the total number of rigid bodies of the manipulator, $N_p = 11$ for 3 piston-rods, one base, one moving platform and 6 rods. $N_t$ is the number of the joints, $N_t = 15$ for 12 spherical joints S, 3 prismatic joints P. $f_i$ denotes the number of degrees of freedom (DOF) of the $i^{th}$ joint, and $m_{\text{int}}$ is the number of internal DOF, which do not influence the motion of manipulator.

Based on equation (1), the mobility of the platform is given by $m = 6(11-15-1) + 39 - 6 = 3$.

Due to the arrangement of the links and joints, as shown in figure 2, legs II and III prevent the platform from rotating about y and z axes. Leg I prevents the platforms from rotating about z-axis but, because $A_{11}A_{12} \neq B_{11}B_{12}$, a slight coupled rotation about x-axis exists.

## III. Kinematic equations

In order to analyse the kinematics of our parallel module, two relative coordinates are assigned as shown in figures 2. A static Cartesian frame xyz is fixed at the base of the machine tool, with the z-axis pointing downward along the vertical direction. The mobile Cartesian frame, $x_P y_P z_P$, is attached to the moving platform at point P and remains parallel to xyz.

In any constrained mechanical system, joints connecting bodies restrict their relative motion and impose constraints on the generalized coordinates, geometric constraints are then formulated as algebraic expressions involving generalized coordinates.

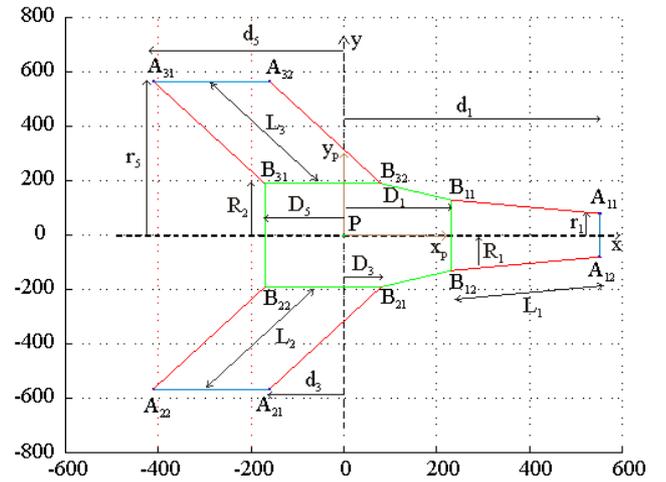

Fig. 4. Dimensions of the parallel kinematic structure in the frame supplied by Fatronik

Using the parameters defined in figure 4, the constraint equations of the parallel manipulator are expressed as:

$$(x_{Bij} - x_{Aij})^2 + (y_{Bij} - y_{Aij})^2 + (z_{Bij} - z_{Aij})^2 - L_i^2 = 0 \quad (2)$$

where $A_{ij}$ (respectively $B_{ij}$) is the center of spherical joint number j on the prismatic joint number i (respectively on the moving platform side), i = 1..3, j = 1..2.

Leg I is represented by two different equations (3) and (4). This is due to the fact that $A_{11}A_{12} \neq B_{11}B_{12}$ (figure 4).

$$(x_P + D_1 - d_1)^2 + (y_P + R_1\cos(\alpha) - r_1)^2 \\ + (z_P + R_1\sin(\alpha) - \rho_1)^2 - L_1^2 = 0 \quad (3)$$

$$(x_P + D_1 - d_1)^2 + (y_P - R_1\cos(\alpha) + r_1)^2 \\ + (z_P - R_1\sin(\alpha) - \rho_1)^2 - L_1^2 = 0 \quad (4)$$

Leg II is represented by a single equations (5).

$$(x_P + D_2 - d_2)^2 + (y_P - R_2\cos(\alpha) + r_4)^2 \\ + (z_P - R_2\sin(\alpha) - \rho_2)^2 - L_2^2 = 0 \quad (5)$$

The leg III, which is similar to leg II (figure 4), is also represented by a single equation (6).

$$(x_P + D_2 - d_2)^2 + (y_P + R_2\cos(\alpha) - r_4)^2 \\ + (z_P + R_2\sin(\alpha) - \rho_3)^2 - L_3^2 = 0 \quad (6)$$

## IV. Coupling between the position and the orientation of the platform

The parallel module of the VERNE machine possesses three actuators and three degrees of freedom. However, there is a coupling between the position and the orientation angle of the platform. The object of this section is to study the coupling constraint imposed by leg I.

By eliminating $\rho_1$ from equations (3) and (4), we obtain a relation (7) between $x_P$, $y_P$ and $\alpha$ independently of $z_P$.

$$R_1^2\sin(\alpha)^2(x_P + D_1 - d_1)^2 + (r_1^2 - 2R_1 r_1\cos(\alpha) + R_1^2)y_P^2 \\ - R_1^2\sin(\alpha)^2(L_1^2 - (R_1^2 + r_1^2 - 2R_1 r_1\cos(\alpha))) = 0 \quad (7)$$

We notice that for a given $\alpha$, equation (7) represents an ellipse (8). The size of this ellipse is determined by $a$ and $b$, where $a$ is the length of the semi major axis and $b$ is the length of the semi minor axis.

$$\frac{(x_P + D_1 - d_1)^2}{a^2} + \frac{y_P^2}{b^2} = 1 \quad (8)$$

where
$$\begin{cases} a = \sqrt{(L_1^2 - (R_1^2 + r_1^2 - 2R_1 r_1\cos(\alpha)))} \\ b = \sqrt{\dfrac{R_1^2\sin(\alpha)^2(L_1^2 - (R_1^2 + r_1^2 - 2R_1 r_1\cos(\alpha)))}{(r_1^2 - 2R_1 r_1\cos(\alpha) + R_1^2)}} \end{cases}$$

These ellipses define the locus of points reachable with the same orientation $\alpha$.

## V. The Inverse kinematics

The inverse kinematics deals with the determination of the joint coordinates as function of the moving platform position. For the inverse kinematics problem of our spatial parallel manipulator, the position coordinates ($x_P$, $y_P$, $z_P$) are given but the joint coordinates $\rho_i$ ($i = 1..3$) of the actuated prismatic and the orientation angle $\alpha$ of the moving platform are unknown.

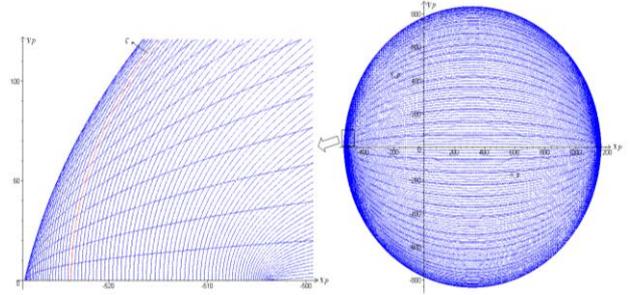

Fig. 5. Curves of iso-values of the orientation $\alpha$ from $-\pi$ to $+\pi$ following a constant step of $\pi/90$.

To solve the inverse kinematics problem, we first find all the possible orientation angles $\alpha$ for prescribed values of the position of the platform ($x_P$, $y_P$, $z_P$). These orientations are determined by solving equation (9), a third degree characteristic polynomial in $\cos(\alpha)$ derived from equation (7).

$$p_1\cos(\alpha)^3 + p_2\cos(\alpha)^2 + p_3\cos(\alpha) + p_4 \quad (9)$$

where
$$\begin{cases} p_1 = 2R_1^3 r_1, \ p_3 = -2R_1^3 r_1 - 2R_1 r_1 y_P^2 \\ p_2 = R_1^2(L_1^2 - R_1^2 - r_1^2) - R_1^2(x_P + D_1 - d_1)^2 \\ p_4 = R_1^2(x_P + D_1 - d_1)^2 + (R_1^2 + r_1^2)y_P^2 \\ \quad - R_1^2(L_1^2 - R_1^2 - r_1^2) \end{cases}$$

However this equation also represents ellipses of iso-values of $\alpha$. So if we plot all ellipses together by varying $\alpha$ from $-\pi$ to $+\pi$ (figure 5), we notice that every point (defined by $x_P$, $y_P$ and $z_P$) is obtained by the intersection of two ellipses and each ellipse represent two opposite orientations so each point can have a maximum of four different orientations. This conclusion is verified by the fact that we can only find four real solutions to the polynomial (Table I).

| $\begin{cases} x_P, y_P, z_P \\ y_P \neq 0 \end{cases}$ | $\alpha = \{\pm\alpha_1 \text{ and } \pm\alpha_2\}$ |
|---|---|
| $\begin{cases} x_P, y_P, z_P \\ y_P = 0 \end{cases}$ | $\alpha = \{0, \pm\alpha_1, \pi\}$ |

TABLE I. the possible orientations for a fixed position of the platform

After finding all the possible orientations, we use the equations derived in section III to calculate the joint coordinates $\rho_i$ for each orientation angle $\alpha$. To make this task easier, we introduce two new points $A_1$ and $B_1$ as the middle of $A_{11}A_{12}$ and $B_{11}B_{12}$, respectively.

$$(x_P + D_1 - d_1)^2 + y_P^2 + (z_P - \rho_1)^2 \\ -(L_1^2 - (R_1^2 + r_1^2 - 2R_1 r_1 \cos(\alpha))) = 0 \quad (10)$$

Then, for prescribed values of the position and orientation of the platform, the required actuator inputs can be directly computed from equations (10), (5) and (6):

$$\rho_1 = z_P + s_1 \sqrt{\begin{array}{c}(L_1^2 - (R_1^2 + r_1^2 - 2R_1 r_1 \cos(\alpha))) \\ -(x_P + D_1 - d_1)^2 - y_P^2\end{array}} \quad (11)$$

$$\rho_2 = z_P - R_2 \sin(\alpha) + s_2 \sqrt{\begin{array}{c}L_2^2 - (x_P + D_2 - d_2)^2 \\ -(y_P - R_2 \cos(\alpha) + r_4)^2\end{array}} \quad (12)$$

$$\rho_3 = z_P + R_2 \sin(\alpha) + s_3 \sqrt{\begin{array}{c}L_3^2 - (x_P + D_2 - d_2)^2 \\ -(y_P + R_2 \cos(\alpha) - r_4)^2\end{array}} \quad (13)$$

where $s_1, s_2, s_3 \in \{\pm 1\}$ are the configuration indices defined as the signs of $\rho_1 - z_P$, $\rho_2 - z_P + R_2 \sin(\alpha)$, $\rho_3 - z_P - R_2 \sin(\alpha)$, respectively.

Subtracting equation (3) from equation (4), yields:
$$y_P (R_1 \cos(\alpha) - r_1) = R_1 \sin(\alpha)(\rho_1 - z_P) \quad (14)$$

$(14) \Rightarrow \text{sgn}(\rho_1 - z_P)\text{sgn}(\sin(\alpha)) = \text{sgn}(R_1 \cos(\alpha) - r_1)\text{sgn}(y_P)$

This means that for prescribed values of the position and orientation of the platform, the joint coordinate $\rho_1$ possesses one solution, except when $\alpha = \{0, \pi\}$. In this case $s_1$ can take on both values +1 and −1. As a result $\rho_1$ can take on two values when $\alpha = \{0, \pi\}$.

| | |
|---|---|
| $\alpha = \{0, \pi\}$ | $s_1 = \pm 1$ |
| $\begin{cases} R_1 \cos(\alpha) = r_1 \\ y_P = 0 \text{ with } \alpha \neq 0 \end{cases}$ | $\rho_1 = z_P$ |
| others | $s_1 = +1$ or -1 |

TABLE II. the solution of the joint coordinate $\rho_1$ according to the values of $\alpha$

Observing equations (11), (12), (13), Table I and Table II, we conclude that the three legs, with four postures for leg I and two postures for leg II and III results in sixteen inverse kinematic solutions (figure 6).

From the sixteen theoretical inverse kinematics solutions shown in figure 6, only one is used by the VERNE machine: the one referred to as (m) in figure 6, which characterized by the fact that each leg must have its slider attachment points upper than the moving platform attachment points, i.e. $s_i = -1$ (remember that the z-axis is directed downward).

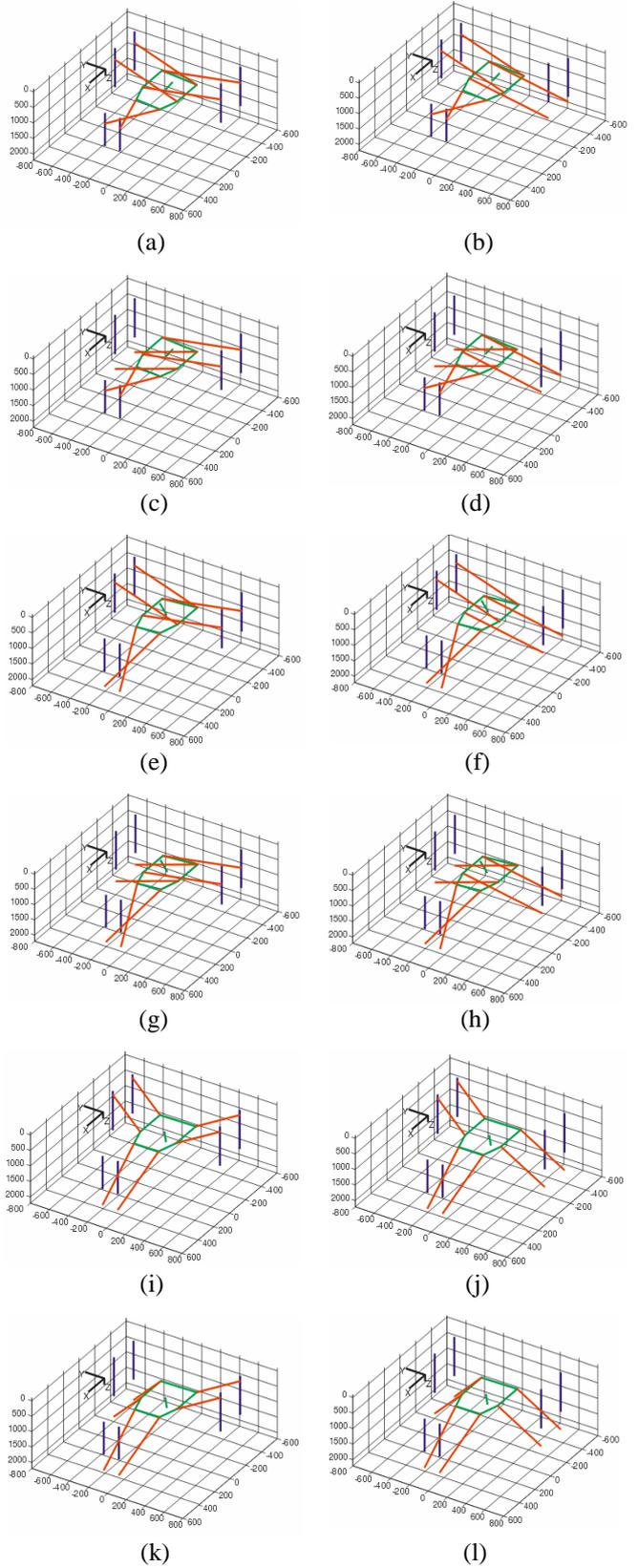

(a) (b) (c) (d) (e) (f) (g) (h) (i) (j) (k) (l)

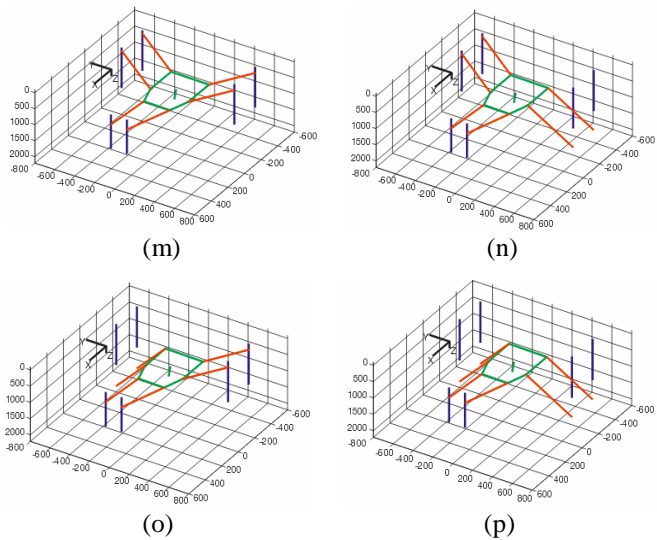

(m)    (n)

(o)    (p)

Fig. 6. the sixteen solutions to the inverse kinematics problem when $x_P = -240$ mm, $y_P = -86$ mm and $z_P = 1000$ mm

For the remaining 15 solutions one of the sliders leave its joint limits, or the two rods of leg I cross. Most of these solutions are characterized by the fact that at least one of the legs has its slider attachment points lower than the moving platform attachment points. So only $s_1, s_2, s_3 = -1$ in equations (11), (12) and (13) must be selected (remember that the z-axis is directed downward). To prevent rod crossing, we also add a condition on the orientation of the moving platform. This condition is $R_1 \cos(\alpha) > r_1$. Finally, we check the joint limits of the sliders and the serial singularities [16].

For the VERNE parallel module, applying the above conditions will always yield to a unique solution for practical applications (solution (m) shown in figure 6).

## VI. The forward kinematics

The forward kinematics deals with the determination of the moving platform position as function of the joint coordinates. For the forward kinematics of our spatial parallel manipulator, the values of the joint coordinates $\rho_i$ ($i=1..3$) are known and the goal is to find the coordinates $x_P$, $y_P$ and $z_P$ of the tool centre point P

To solve the forward kinematics, we successively eliminate variables $x_P$, $y_P$ and $z_P$ from the system ($S1$) of four equations ((3), (4), (5) and (6)) to lead to an equation function of the joint coordinates $\rho_i$ ($i=1..3$) and function of the orientation angle $\alpha$ of the platform. To do so, we first compute $y_P$ as function of $z_P$ by subtracting equation (3) from equation (4) and we replace this variable in system ($S1$) to obtain a new system ($S2$) of three equations (15), (16) and (17) derived from equations (3), (5) and (6) respectively. We then compute $z_P$ as function of $\rho_i$ ($i=1..3$) and $\alpha$ by subtracting equation (16) from equation (17). We replace this variable in system ($S2$) to obtain a new system ($S3$) of two equations (18) and (19) derived from equations (15) and (16) respectively. Finally, we compute $x_P$ as function of $\rho_i$ ($i=1..3$) and $\alpha$ by subtracting equation (18) from equation (19) and we replace this variable in the system ($S3$) in order to eliminate $x_P$.

Equations of system ($Si$) (i=2..3) are not reported here because of space limitation. They are available in [16].

For each step, we determine solutions existence conditions by studying the denominators that appear in the expressions of $x_P$, $y_P$ and $z_P$. These conditions are:

$$R_1 \cos(\alpha) - r_1 \neq 0 \quad (20)$$

$$(\rho_2 - \rho_3)(R_1 \cos(\alpha) - r_1) + 2\sin(\alpha)(r_4 R_1 - r_1 R_2) \neq 0 \quad (21)$$

Equation (20) obtained from (14) implies that $A_1 B_1$ is perpendicular to the slider plane of leg I. In this case equation (8) represents a circle because $a = b$.

When $\rho_2 = \rho_3$ in equation (21), we have $\alpha = \{0, \pi\}$. This means that $y_P = 0$ (obtained from Equations. (5) – (6)).

To finish the resolution of the system, we perform the tangent-half-angle substitution $t = \tan(\alpha/2)$. As a consequence, the forward kinematics of our parallel manipulator results in a eight degree characteristic polynomial in $t$, whose coefficients are relatively large expressions in $\rho_1$, $\rho_2$ and $\rho_3$. For the VERNE machine, only 4 assembly-modes have been found (figure 7). It was possible to find up to 6 assembly-modes but only for input joint values out of the reachable joint space of the machine.

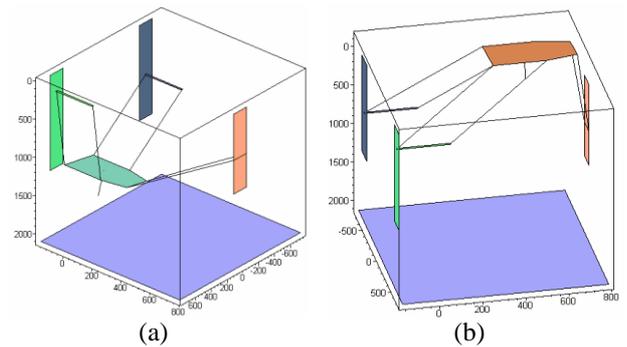

(a)    (b)

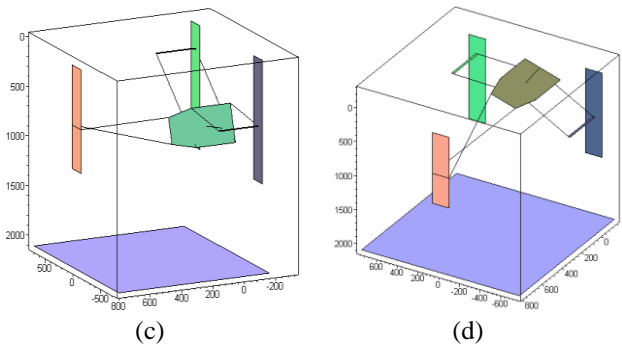

Fig. 7. The four assembly-modes of the VERNE parallel module for $\rho_1 = 250$ mm, $\rho_2 = 1000$ mm and $\rho_3 = 750$ mm

Only one assembly-mode is actually reachable by the machine (solution (a) shown in figure 7) because the other ones lead to either rod crossing, collisions, or joint limit violation. The right assembly mode can be recognized, like for the right working mode, by the fact that each leg must have its slider attachment points upper than the moving platform attachment points, i.e. $s_i = -1$ (keep in mind that the z-axis is directed downwards).

The proposed method for calculating the various solutions of the forward kinematic problem has been implemented in Maple (Table III).

| $\rho_1 = 674$ mm, $\rho_2 = 685$ mm and $\rho_3 = 250$ mm | | | | |
|---|---|---|---|---|
| Case | $t$ (rd) | $x_P$ (mm) | $y_P$ (mm) | $z_P$ (mm) |
| (a) | -0.22 | -199.80 | 355.92 | 1242 |
| (b) | -0.14 | 298.35 | -297.53 | -120.22 |
| (c) | 1.81 | -393.6 | 322.82 | 958.21 |
| (d) | 2.70 | -115.62 | -189.68 | -0.26 |

TABLE III. the numerical results of the forward kinematic problem of the example where $\rho_1 = 674$ mm, $\rho_2 = 685$ mm and $\rho_3 = 250$ mm

## VII. Conclusion

This paper was devoted to the kinematic analysis of the parallel module of a 5-DOF hybrid machine tool, the VERNE machine. The degrees of freedom, the inverse kinematics and the different assembly modes were derived. The forward kinematics was solved with the substitution method. It was shown that the inverse kinematics has sixteen solutions and the forward kinematics may have six real solutions. Examples were provided to illustrate the results. The forward and inverse kinematics of the full VERNE machine is quite easy to derive [16]. The controller of the actual VERNE machine resorts to an iterative Newton-Raphson resolution of the kinematics models. A comparative study will be conducted by the authors between the analytical and the iterative approaches. It is expected that the analytical method could decrease the Cpu-time and improve the quality of the control.

**Acknowledgments**

This work has been partially funded by the European projects NEXT, acronyms for "Next Generation of Productions Systems", Project no° IP 011815.